\documentclass[sigconf]{acmart}

\usepackage{xstring}
\usepackage{listings}
\usepackage{color}

\usepackage{balance}

\definecolor{darkgray}{gray}{0.3}
\definecolor{lightgray}{gray}{0.95}

\renewcommand\footnotetextcopyrightpermission[1]{} % removes footnote with conference information in first column
\setcopyright{none}
\title{Deep Learning with Apache SystemML}
\author{Niketan Pansare$^1$, Michael Dusenberry$^2$, Nakul Jindal$^2$, \\
	Matthias Boehm$^1$, Berthold Reinwald$^1$, Prithviraj Sen$^1$}
\affiliation{$^1$IBM Research - Almaden \\
	$^2$IBM Spark Technology Center}

\begin{document}

\maketitle

%\begin{abstract}
%TODO abstract
%\end{abstract}

\section{Introduction}
Deep Learning (DL) is a subfield of Machine Learning (ML) that focuses 
on learning hierarchical representations of data with multiple levels of abstraction
using neural networks~\cite{dlNatureArticle}. 
Recent advances in deep learning are made possible due to the availability of large amounts of labeled data, use of GPGPU compute, and 
application of new techniques (such as ReLU, batch normalization~\cite{DBLP:journals/corr/IoffeS15}, dropout~\cite{JMLR:v15:srivastava14a}, residual block~\cite{DBLP:journals/corr/HeZRS15}, etc.) that help deal with issues in training deep networks.
In spite of the need to train on large datasets, there is a disconnect between the deep learning community and the big data community. 
To scale to a multi-node cluster, most deep learning frameworks (such as Caffe2, TensorFlow~\cite{Abadi:2016:TSL:3026877.3026899} and IBM's PowerAI DDL~\cite{DBLP:journals/corr/abs-1708-02188}) 
use custom communication libraries based on either MPI (such as IBM Spectrum MPI, Facebook's Gloo) or a custom networking protocol (such as Google RPC).
Unlike popular big data frameworks (such as Apache Hadoop~\cite{mapreduce-osdi} and Apache Spark~\cite{Zaharia:2010:SCC:1863103.1863113}), these communication libraries do not provide features such as resource sharing, multi-tenancy and fault-tolerance out of the box, 
making them difficult to deploy on shared production clusters. 
This leads to ineffective use of resources in an organization, often requiring two separate infrastructures (i.e. scale-up versus scale-out).
This problem is even more severe when the data generated as part of the big data pipeline 
(ML, data preprocessing, data cleaning) needs to be consumed by the deep learning pipeline or vice versa,
as the workload characteristics of a typical machine learning algorithm (i.e. memory-bound, BLAS level-2, sparse/ultra-sparse inputs (or feature matrix), etc.) are often
different than that of a typical deep learning algorithm (i.e. compute-bound, BLAS level-3, dense inputs, etc.).
Apache SystemML~\cite{Boehm:2016:SDM:3007263.3007279} aims to bridge that gap
by seamlessly integrating with underlying big data frameworks
and by providing a unified framework for implementing machine learning and deep learning algorithms.

In Apache SystemML, the ML algorithms are implemented using a
high-level R-like language called DML (short for Declarative Machine Learning).
DML improves the productivity of data scientists by enabling them to 
implement their ML algorithm with precise semantics as well as abstract data types
and operations, independent of the underlying data representation or cluster characteristics.
For the given DML script, SystemML's cost-based compiler automatically generates hybrid runtime
execution plans that are composed of single-node and distributed
operations depending on data and cluster characteristics
such as data size, data sparsity, cluster size and memory
configurations, while exploiting the capabilities of underlying
data-parallel frameworks such as MapReduce or Spark.
This allows for algorithm reusability across data-parallel frameworks, and simplified deployment for varying data characteristics
and runtime environments, ranging from low-latency scoring to large-scale training.

%\textbf{TODO:} Introduce Deep Learning 

\section{Deep Learning APIs}

\textbf{NN Library.} As with other ML algorithms, users can implement their deep learning models using DML. 
SystemML 1.0 does not support automatic differentiation, thus the user has to write the DML code for the 
partial derivatives (i.e., the backward pass) of each layer. 
SystemML ships with a \texttt{Neural Network} (NN) library 
that supports 20+ pre-implemented layers (for example: conv2d, affine, relu, etc.) 
and 6 optimizers (namely Adagrad, Adam, RMSprop, SGD, SGD with momentum, and SGD with Nesterov momentum) to assist in writing algorithms. 
Each layer in the NN library has an \texttt{init}, \texttt{forward}, and \texttt{backward} function. 
%Each layer in the NN library has a \texttt{forward} and \texttt{backward} function which can be used 
%without any additional modification in the forward and the backward pass of training algorithm.
%Additionally, some layers (such as conv2d, affine, etc.) have an optional \texttt{init} function
%that can be used to initialize the weights and bias.
The NN library is implemented entirely in DML, allowing the user to 
conveniently add custom layers and modify existing layers.
The DML script for training a softmax classifier in SystemML
using the minibatch gradient descent algorithm
and the affine, softmax, and cross entropy layers 
is given below:

% numbers=left
%\begin{lstlisting}[language=R, basicstyle=\tiny]
\begin{lstlisting}[frame=tb, language=R]
source("nn/layers/affine.dml") as affine
source("nn/layers/cross_entropy_loss.dml") as cross_entropy_loss
source("nn/layers/softmax.dml") as softmax
source("nn/optim/sgd.dml") as sgd
train = function(matrix[double] X, matrix[double] Y) {
  D = ncol(X)  # num features
  K = ncol(Y)  # num classes
  lr = 0.01; batch_size = 32; num_iter = nrow(X) / batch_size 
  [W, b] = affine::init(D, K)
  for(i in 1:num_iter) {
    # Get batch
    beg = (i-1)*batch_size + 1; end = beg + batch_size
    X_batch = X[beg:end,]; y_batch = Y[beg:end,]
    # Perform forward pass
    scores = affine::forward(X_batch, W, b) # or X_batch %*% W + b
    probs = softmax::forward(scores)
    # Perform backward pass
    dprobs = cross_entropy_loss::backward(probs, y_batch)
    dscores = softmax::backward(dprobs, scores)
    [dX_batch, dW, db] = affine::backward(dout, X_batch, W, b)
    # Perform update
    W = sgd::update(W, dW, lr)
    b = sgd::update(b, db, lr)
  }
}
\end{lstlisting}

\textbf{Keras2DML/Caffe2DML API.} 
Currently active DL experts may be familiar with popular packages such as Keras~\cite{chollet2015keras} or Caffe~\cite{jia2014caffe} 
and may want to avoid learning a new language like DML.
Or one may want to use publicly available pre-trained Keras/Caffe models
and also leverage SystemML's distributed execution on Spark.
To support such users, SystemML ships with python APIs - 
\texttt{Keras2DML} and \texttt{Caffe2DML} - 
that accept the DL models expressed in Keras or Caffe format and
generate the equivalent DML script. 
Furthermore, these APIs allow a Python programmer 
to invoke SystemML's algorithms using a \texttt{scikit-learn} like API (which accepts NumPy arrays, SciPy matrices, or Pandas DataFrames)
as well as Spark's MLPipeline API (which accepts \texttt{Spark DataFrame}s).
Since these APIs conform to MLPipeline's Estimator interface,
they can be used in tandem with MLLib's feature extractors, transformers, scoring 
and cross-validation classes. 
Also, these APIs support loading pre-trained weights in the Caffe and Keras format
for transfer learning and prediction. 
Assuming input matrices \texttt{X} and \texttt{Y} are NumPy arrays,
the equivalent Python code for the above DML script is as follows:

\begin{lstlisting}[frame=tb, language=Python]
from keras.models import Sequential
from keras.layers import Dense
from keras.optimizers import SGD
from systemml.mllearn import Keras2DML
def train(X, Y):
  num_elems, D = X.shape; K = Y.shape[1]
  epochs = num_iter / num_elems
  model = Sequential()
  model.add(Dense(K, activation="softmax", input_dim=D))
  sgd = SGD(lr=0.01, momentum=0, nesterov=False)
  model.compile(loss="categorical_crossentropy", optimizer="sgd")
  sysml_model = Keras2DML(spark, model, input_shape=(D,1,1))
  sysml_model.set(train_algo="minibatch", test_algo="allreduce")
  sysml_model.fit(X, Y)
\end{lstlisting}
% sysml_model.fit(X, Y, batch_size=batch_size, epochs=epochs)
% sysml_model.set(train_algo="minibatch") # default

\section{Compiler and Runtime Support}
 
\textbf{Tensor Representation.}
The primary data structure to store large amounts of data in DML is a 2D Matrix whereas in typical DL applications, a multi-dimensional matrix or a tensor is commonly used.
To represent a tensor with more than 2 dimensions in DML, we linearize all but the first dimension. Therefore,
a 4-dimensional tensor of shape $[ N, C, H, W ]$ is represented as a matrix
with $N$ rows and $C*H*W$ columns. 
This simplification helps us to leverage existing physical optimizations such as various sparse formats  
(COO, CSR and Modified CSR), blocking for handling out-of-core tensors, 
and broadcasting operations over scalars and vectors.
Other optimizations such as sum-product optimization and
code generation are also leveraged when applicable.

\textbf{Builtin NN Functions.} 
SystemML provides a set of built-in functions that are part of the language. Some common ones are \texttt{min}, \texttt{max}, \texttt{mean}, \texttt{sum}, and \texttt{solve}.
Even though convolution and pooling (and their respective backward functions) 
can be expressed using existing DML looping constructs (as they are in the NN library),
we've added them as built-in functions to enable efficient implementations. 
In addition to supporting ML use cases in SystemML, using the aforementioned representation of tensors and  built-in functions, we support
a variety of deep learning models in SystemML such as LeNet, feedforward nets, ResNets, autoencoders, simple RNNs, LSTMs, U-Net, 
SentenceCNN, etc.
%Since th above simplifications are not SystemML-specific, 
%we claim that the existing linear algebra frameworks/libraries can be extended using 
%the above simplifications to support deep learning without loss of generality. % or efficiency.
%Like SystemML, most deep learning frameworks cast tensor operations such as convolution as a matrix multiplication to exploit the underlying BLAS. 

\textbf{Native BLAS Exploitation.} To exploit the low-level CPU SIMD instructions, we extended the SystemML runtime to
use the underlying BLAS (such as OpenBLAS and Intel MKL) for compute-intensive operations such as matrix-matrix multiplication and
convolution operations. If Intel MKL is installed, the SystemML runtime uses the highly-tuned MKL-DNN primitives for the convolution operations.

\textbf{GPU Backend.} Most of the time in training a deep neural network is
spent in matrix multiplications and convolution operations~\cite{Jia:EECS-2014-93}.
These operations can be performed, in case of dense inputs or intermediates, extremely efficiently on a GPU
which often leads to a speedup of 10x as compared to CPU.
Hence, it became imperative to support GPU backend in SystemML.
To add a GPU backend, the SystemML optimizer was 
modified to compile a GPU low-level operator
if the input data, intermediate data and output data 
for a given operation fits in the GPU device memory.
The GPU backend invokes highly tuned kernels from CUDA libraries like CuBLAS, CuSPARSE, or CuDNN when available. In other instances, it invokes custom CUDA kernels packaged with SystemML.
Data is lazily copied back and forth between the GPU device memory and the host memory as needed.
Also, data is converted from row-major to column-major and vice-versa when needed by CUDA library operations. 
%Additionally, we added a ``GPU buffer pool'' that supports the conversion 
%of various matrix formats (for example: MCSR/COO to CSR, binary blocked RDD to row-major dense, etc.)
%as well as the transfer of matrices from other backends (such as Spark, single-node) to GPU device memory.
%To support efficient training of large models,
%the GPU buffer pool is  designed to minimize
%the CPU-GPU transfers and is
%backed by an in-memory and disk-based bufferpool 
%on the driver. This allows for evictions of unpinned matrices
%from device to host (using a LRU strategy) when the GPU runs out of memory.
Data is evicted from the GPU memory using an LRU strategy. It is copied back to the host memory if it was dirty when evicted. Data on the host is spilled onto disk when appropriate.
%Hence, %unlike other frameworks,
%there is no restriction on the depth or size of the deep learning model. 
%and the user need not worry about deployment related issues such as
%device memory size when designing the neural network.
%In the above example, the script can be GPU-enabled using the  \texttt{sysml\_model.setGPU(True)} method.

\textbf{Sparse Operations.} SystemML maintains the 
number of non-zeros for each intermediate matrix,
decides upon dense or sparse formats,
and selects appropriate runtime operators for 
combinations of dense and sparse inputs.
For sparse-safe operations (such as convolution and matrix multiplication), 
this reduces the number of floating point operations and 
improves memory efficiency.
%amount of memory required for the given operation
%and hence leads to faster end-to-end performance.
For example, there are four physical convolution operators
(using lowering technique~\cite{DBLP:journals/corr/ChetlurWVCTCS14}),
dense input / dense filter, sparse input / dense filter,  dense input / sparse filter
and sparse input / sparse filter. 

\textbf{Distributed Operations.} 
The \texttt{Keras2DML} and \texttt{Caffe2DML} APIs
allow the user to configure the execution strategy of the underlying optimization
algorithm with the parameters \texttt{train\_algo} and \texttt{test\_algo}.
For example, if \texttt{train\_algo} is set to \texttt{"minibatch"}, 
the generated DML script % (similar to the script in the previous section) 
contains a \texttt{for} loop
that loops over the dataset one batch at a time. 
%By setting the \texttt{test\_algo} parameter to \texttt{"allreduce"}, the parallel loop construct - \texttt{parfor} - is included in the generated DML. 
%by setting the \texttt{test\_algo} parameter to \texttt{"allreduce"}. %the appropriate parameter.
%and use SystemML's parallel loop construct to compute the loss over the validation dataset and predicting the labels over the test dataset. The parallel loop construct is implemented through threads on a single machine or Map-Reduce on a cluster.
If \texttt{batch\_size} is small enough 
such that the input, output and intermediate matrices fit in the driver JVM, 
then SystemML will generate a single-node plan.
If on the other hand, the user sets \texttt{batch\_size} to a very large value 
(for example, \texttt{train\_algo} set to \texttt{"batch"}) or
if the weights no longer fit on the driver JVM,
then SystemML will generate a distributed data-parallel plan
where the large input activations or weights are partitioned into fixed size blocks and 
represented internally as \texttt{RDD}~\cite{Boehm:2016:SDM:3007263.3007279}. 
%The low-level runtime operators on these matrices are implemented using Spark's APIs.
For scoring using a compute-intensive deep network such as ResNet-50 on a large dataset,
it is often better to use the task-parallel loop construct - \texttt{parfor} - 
with a small \texttt{batch\_size} instead of the \texttt{for} loop~\cite{Boehm:2014:HPS}.
%In this scenario, SystemML's optimizer parallelizes based on 
This script is generated automatically by setting \texttt{test\_algo} to \texttt{"allreduce"}.
%As described previously, when using the \texttt{Keras2DML} and \texttt{Caffe2DML} APIs, the parallel loop construct - \texttt{parfor} - can be included in the generated DML by setting the \texttt{test\_algo} parameter to \texttt{"allreduce"}. %the appropriate parameter.
The \texttt{parfor} optimizer then automatically creates optimal parallel 
execution plans that exploits multi-core, multi-gpu, and cluster parallelism
based on the underlying cluster and data characteristics.
As an example, the \texttt{parfor} optimizer compiles a 
row-partitioned \texttt{remote-parfor} plan for the ResNet-50 prediction script that avoids shuffling
and scales linearly with the number of cluster nodes over large data.

\section{Future Work} We plan
to extend the existing code generation framework in SystemML 
for deep learning operations. This includes supporting
vertical fusion across layers, horizontal fusion for shared inputs
(e.g. reuse temporary \texttt{im2col} intermediates 
in presence of multiple convolution operators consuming the same input), 
and code generation for heterogeneous hardware including GPUs.
%For example, reusing intermediate lowered matrix 
%leads to significant speedup especially if
%the input batch is sparse and is consumed
%by multiple conv2d operators. ( \textbf{model-parallel})
Asynchronous algorithms 
such as HogWild!~\cite{HogWild}, and Stale-Synchronous SGD~\cite{Ho2013} 
will be supported in SystemML through parameter server abstractions~\cite{systemml2083}.
This will help in making SystemML a unified framework for small- and large-scale machine learning 
that supports data-parallel, task-parallel, and parameter-server-based execution strategies in a single framework.
We also plan to investigate the automatic exploitation
of the optimization tradeoff between hardware efficiency 
and statistical efficiency~\cite{DeSa:2017:UOA:3140659.3080248}. % on novel hardware.
%Also, we plan to add primitives to
%support algorithms such as low-precision/asynchronous SGD
%and evaluate the optimization tradeoff between hardware efficiency 
%and statistical efficiency. % on novel hardware.
% as the im2col consumes a significant portion of the execution time
%We also plan to add auto-differentiation primitives in SystemML 
%to allow the end user to train a deep learning network 
%without implementing the DML-bodied backward function for custom layers.

%\newpage
\balance
\bibliography{main} 

\begin{thebibliography}{10}

\bibitem{systemml2083}
{SYSTEMML-2083 - Language and Runtime for Parameter Servers}.
\newblock \url{https://issues.apache.org/jira/browse/SYSTEMML-2083}, 2018.

\bibitem{Abadi:2016:TSL:3026877.3026899}
Mart\'{\i}n Abadi et~al.
\newblock {TensorFlow: A System for Large-scale Machine Learning}.
\newblock In {\em OSDI}, 2016.

\bibitem{Boehm:2014:HPS}
Matthias Boehm et~al.
\newblock {Hybrid Parallelization Strategies for Large-scale Machine Learning
  in SystemML}.
\newblock {\em PVLDB}, 7(7), 2014.

\bibitem{Boehm:2016:SDM:3007263.3007279}
Matthias Boehm et~al.
\newblock {SystemML: Declarative Machine Learning on Spark}.
\newblock {\em PVLDB}, 9(13), 2016.

\bibitem{DBLP:journals/corr/ChetlurWVCTCS14}
Sharan Chetlur et~al.
\newblock {cuDNN: Efficient Primitives for Deep Learning}.
\newblock {\em CoRR}, 2014.

\bibitem{DBLP:journals/corr/abs-1708-02188}
Minsik Cho et~al.
\newblock {PowerAI DDL}.
\newblock {\em CoRR}, 2017.

\bibitem{chollet2015keras}
Fran\c{c}ois Chollet et~al.
\newblock Keras.
\newblock \url{https://github.com/fchollet/keras}, 2015.

\bibitem{DeSa:2017:UOA:3140659.3080248}
Christopher De~Sa, Matthew Feldman, Christopher R{\'e}, and Kunle Olukotun.
\newblock {Understanding and Optimizing Asynchronous Low-Precision Stochastic
  Gradient Descent}.
\newblock {\em SIGARCH}, 45, 2017.

\bibitem{mapreduce-osdi}
Jeff Dean et~al.
\newblock {MapReduce: Simplified Data Processing on Large Clusters}.
\newblock In {\em OSDI}, 2004.

\bibitem{DBLP:journals/corr/HeZRS15}
Kaiming He et~al.
\newblock {Deep Residual Learning for Image Recognition}.
\newblock {\em CoRR}, 2015.

\bibitem{Ho2013}
Qirong Ho et~al.
\newblock {More Effective Distributed ML via a Stale Synchronous Parallel
  Parameter Server}.
\newblock In {\em NIPS}, 2013.

\bibitem{DBLP:journals/corr/IoffeS15}
Sergey Ioffe et~al.
\newblock {Batch Normalization: Accelerating Deep Network Training by Reducing
  Internal Covariate Shift}.
\newblock {\em CoRR}, 2015.

\bibitem{Jia:EECS-2014-93}
Yangqing Jia.
\newblock {\em {Learning Semantic Image Representations at a Large Scale}}.
\newblock PhD thesis, EECS Department, University of California, Berkeley, May
  2014.

\bibitem{jia2014caffe}
Yangqing Jia et~al.
\newblock {Caffe: Convolutional Architecture for Fast Feature Embedding}.
\newblock {\em CoRR}, 2014.

\bibitem{dlNatureArticle}
Yann LeCun et~al.
\newblock {Deep Learning}.
\newblock {\em Nature}, 521, 2015.

\bibitem{HogWild}
Benjamin Recht et~al.
\newblock {Hogwild: A Lock-Free Approach to Parallelizing Stochastic Gradient
  Descent}.
\newblock In {\em NIPS}. 2011.

\bibitem{JMLR:v15:srivastava14a}
Nitish Srivastava et~al.
\newblock {Dropout: A Simple Way to Prevent Neural Networks from Overfitting}.
\newblock {\em JMLR}, 2014.

\bibitem{Zaharia:2010:SCC:1863103.1863113}
Matei Zaharia et~al.
\newblock {Spark: Cluster Computing with Working Sets}.
\newblock HotCloud, 2010.

\end{thebibliography}
\bibliographystyle{plain} %ACM-Reference-Format}

\end{document}